\PassOptionsToPackage{dvipsnames}{xcolor}

\documentclass{ngsm2024} % Include author names
\usepackage{microtype}
\usepackage{booktabs}

\usepackage{graphicx}
\usepackage{hyperref}
\usepackage{mathtools}
\usepackage{multirow}
\usepackage{enumitem}
\usepackage{float}
\usepackage{comment}
\usepackage[capitalize, nameinlink, noabbrev]{cleveref}

\usepackage{layouts}

\hypersetup{
    colorlinks,
    linkcolor=NavyBlue,
    citecolor=PineGreen,
    filecolor=Mulberry,
    urlcolor=NavyBlue,
    menucolor=BrickRed,
    runcolor=Mulberry
}

\DeclarePairedDelimiter\floor{\lfloor}{\rfloor}

\title[Towards DFA on Medical Time Series by Maximizing CMI]{Towards Dynamic Feature Acquisition on Medical Time Series\\by Maximizing Conditional Mutual Information}

\optauthor{%
    \Name{Fedor Sergeev\textsuperscript{1}} \Email{fedor.sergeev@inf.ethz.ch}\\
    \Name{Paola Malsot\textsuperscript{1,2}} \Email{paola.malsot@ai.ethz.ch}\\
    \Name{Gunnar Rätsch\textsuperscript{1}} \Email{raetsch@inf.ethz.ch}\\
    \Name{Vincent Fortuin\textsuperscript{3,4,5}} \Email{vincent.fortuin@helmholtz-munich.de}\smallskip\\
    \addr \textsuperscript{1} ETH Zurich, Zurich, Switzerland \\
    \addr \textsuperscript{2} ETH AI Center, ETH Zurich, Switzerland \\
    \addr \textsuperscript{3} Helmholtz AI, Munich, Germany \\
    \addr \textsuperscript{4} Technical University of Munich, Germany \\
    \addr \textsuperscript{5} Munich Center for Machine Learning, Germany
}

\begin{document}
%textwidth in cm: \printinunitsof{in}\prntlen{\linewidth}

\maketitle
\begin{abstract}%
    Knowing which features of a multivariate time series to measure and when is a key task in medicine, wearables, and robotics. Better acquisition policies can reduce costs while maintaining or even improving the performance of downstream predictors. Inspired by the maximization of conditional mutual information, we propose an approach to train acquirers end-to-end using only the downstream loss. We show that our method outperforms random acquisition policy, matches a model with an unrestrained budget, but does not yet overtake a static acquisition strategy. We highlight the assumptions and outline avenues for future work.
\end{abstract}

%\begin{keywords}%
%  List of keywords%
%\end{keywords}

\section{Introduction}
\label{submission}
 
In the medical setting, clinicians often need to monitor patients over time during their hospital stay, especially in Intensive Care Units \citep[ICUs;][]{hyland_early_2020}. They try to improve the patient's state by administering drugs while relying on continuous measurements of vital signs (e.g., heart rate) and occasional lab tests (e.g., blood tests, X-rays). While the continuous measurements are automatic and practically free, performing lab tests takes the clinical staff's time and incurs additional costs. We aim to develop a method for recommending which lab tests to perform, in order to best monitor the patient's state, while decreasing workload and costs.

More formally, the hospital stay of a patient $i$ can be represented as a multivariate (or even multi-modal) time series $\pmb{x}^i=\{x^i_{t,f}\}$ with the features $f$ at time $t$ being the values of either vital signs, lab tests, or administered drugs. Usually, these data are used for time series classification (e.g., mortality prediction), early event prediction (e.g., circulatory failure prediction), or intervention recommendation \citep{hyland_early_2020, yeche_temporal_2023, kuznetsova_importance_2023, liu2020reinforcement}.

\begin{figure}[t]
    \centering
    \begin{minipage}[t]{0.46\textwidth}
        \centering
        \includegraphics[page=1, trim={0 4.4cm 8.65cm 0}, clip, width=\textwidth]{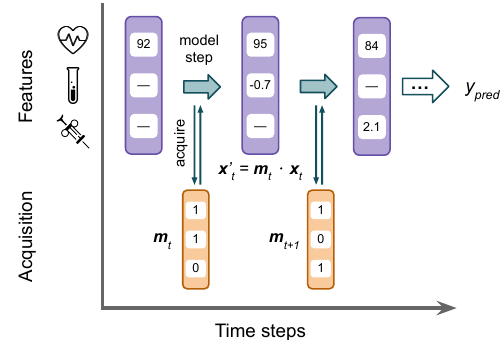}
        \caption{\vspace{-6mm}Sketch of DFA on a regular time series in medicine\protect\footnotemark.}
        \label{fig:dfa}
    \end{minipage}%
    \hfill
    \begin{minipage}[t]{0.46\textwidth}
        \centering
        \includegraphics[page=11,trim={0 4.4cm 8.65cm 0},clip,width=\linewidth]{images/figures.pdf}
        \caption{\vspace{-6mm}Proposed acquisition\ +\ classification mechanism.}
        \label{fig:method}
    \end{minipage}
\end{figure}

We consider the \emph{Dynamic Feature Acquisition} (DFA) task --- based on an observed patient state $\{x^i_{t,f}\}_{t \leq \tau}$ at time $\tau$, recommend which feature(s) $f$ should be measured at some future time $\tau'$ at known cost $c_{\tau,f}$ (see \cref{fig:dfa}). The aim is to reduce the total measurement cost $\sum_{t,f} c_{t,f}$ while maintaining or even improving the performance of a downstream predictor. 

DFA is also relevant for wearables (e.g., extend battery life by reducing the number of sensor activations) \citep{possas2018egocentric, merrill2023homekit2020}, active perception in robotics \citep{bajcsy2018revisiting}, and efficient video classification \citep{yang2022efficient}.\smallskip

Our contributions are:\smallskip
\begin{itemize}[noitemsep, topsep=0pt]
    \item We propose a novel CMI-based approach for DFA. It is compatible with clinically-relevant downstream prediction tasks and can be trained end-to-end.\smallskip
    \item We test on benchmark time series classification datasets with fake features and show that our method outperforms random, matches complete, but falls short of static selection methods.
\end{itemize}

%~\vspace{-18mm}

\footnotetext{Icons: Rockicon, Dilich, Lorc, CC BY 3.0 \href{https://creativecommons.org/licenses/by/3.0}{https://creativecommons.org/licenses/by/3.0}, via Wikimedia Commons.}%
\section{Problem Setting}

Let us assume that the time series are \emph{regular}, indexed with $t \in \{0, \ldots, T^i\}$, where $T^i$ is the length of $\pmb{x}^i$. We consider the case when an acquisition recommendation is made for features that will become available at the next time step (\emph{``next-step'' assumption}): $\tau' = \tau + 1$. For simplicity, we assume that the \emph{measurement cost is constant} over time and features: $c_{t,f} =: c$. Without loss of generality, we set $c = 1$.  Similarly to \citet{kossen_active_2023}, we assume that the \emph{data are fully observed}. 

We set a budget for the total acquisition cost. For static data, the budget is usually be given per sample. For time series, a budget per time step $b(\pmb{x}_t, t)$ should be predicted from the sample budget. In our experiments, we consider a simplified scenario, when it is constant and given a priori: $b(\pmb{x}_t, t) = b$.

The acquisition and prediction cycle under these assumptions is shown in \cref{fig:dfa}. Here, an acquirer is a model that at each time step $\tau$ outputs the \emph{acquisition vector} $\pmb{m}_\tau$. It is a binary vector with ones indicating which features should be acquired at the next time step $\tau + 1$. Since the data are fully observed, we imitate the measurement procedure with an element-wise product. The measured data are then passed to the classifier. Additionally, the acquirer and classifier may have access to each other's internal state. We discuss the assumptions and provide pseudocode of the DFA  cycle in \cref{sec:assumptions}.

Note that the models here are not limited to recurrent architectures: time steps can be accumulated, and the classifier (e.g., a transformer) reapplied  \citep{transformer}. At the same time, by having the classifier receive new data at each time step, we allow for it to be used for both classification and early event prediction (tasks that are relevant for data from ICU and wearables). From this point on, we consider only classification objectives.

\section{Method}

DFA usually follows one of two approaches: use the cost estimate as a penalty function to train the acquisition model using reinforcement learning \citep{kossen_active_2023, yu_deep_2023}, or use some acquisition function to rank and select the most meaningful features \citep{ma_eddi_2019, covert_learning_2023}. The latter approach often uses CMI. Estimating it directly (e.g., using a partial variational autoencoder) can be challenging \citep{ma_eddi_2019}. Instead, CMI can be approximated \citep{covert_learning_2023}. We discuss related work in \cref{sec:related}. 

In \citet{covert_learning_2023}, the authors use a neural network and Categorical distribution to sequentially predict the feature with the largest CMI, and perform (greedy) selection on static data.  Similarly, we use the acquirer neural network to predict logits of the approximate CMI at each time step of the time series (see \cref{fig:method}). We then iteratively (until the budget $b$ is reached) sample a one-hot vector indicating the selected feature using the Gumbel-Softmax \citep[GS;][]{jang2016categorical}. To avoid selecting the same feature twice, we subtract a penalty vector from the acquirer's output. This approach is differentiable and therefore can be trained end-to-end via backpropagation using the classification loss. Further details are available in \cref{sec:appendix}.

\section{Experiments}

We test the proposed method on the \emph{FordA} and \emph{SpokenArabicDigits} datasets from the UCR and UEA time series classification archives \cite{UCRArchive2018, bagnall2018uea}. The data summary and samples are shown in \cref{sec:data}. We consider balanced classification and use accuracy as the performance metric. By default, no features are considered observed; they all have to be explicitly acquired. 

The FordA dataset is univariate, so, to imitate a multivariate dataset, we take $m=10$ consecutive time steps from FordA and set them as one time step with $m$ features of a new \emph{m-FordA} dataset (short for multivariate or multi-step FordA). In contrast, SpokenArabicDigits is multivariate and variable length by design.

\subsection{Fake features}

The features in these datasets are quite similar (i.e., measured by the same device). Therefore it is not obvious whether one feature is more informative than another. To reliably test whether our model learns to acquire the right features, we add $30$ fake features that do not hold any information about the class label (see \cref{fig:normal-fake}). We test three different varieties of fake features: zeros, Gaussian noise, and samples from a Gaussian process (GP). 

We set a constant budget per time step of $b = 5$ and compare our method to a \emph{random} acquisition policy (selects $b$ features at random at each step) and a \emph{complete} acquisition policy (selects all features at each step). This means that the complete acquirer obtains $8$ times more features than the other two. 

The classification accuracy on SpokenArabicDigits is presented in \cref{tab:different-fakes-spoken}, and the acquisition patterns for zeros on m-FordA are presented in \cref{fig:acq-patterns-forda}. Other results, samples, training and implementation details are available in \cref{sec:appendix}.

\begin{table}[h]    
    \centering
    \resizebox{0.475\textwidth}{!}{%
        \begin{tabular}{llccc}
            \hline
            \multirow{2}{*}{Acquirer} & \multicolumn{3}{c}{Type of fake features} \\ \cline{2-4}
            & \multicolumn{1}{l}{Zeros} & \multicolumn{1}{l}{Noise} & GP \\ \hline
            Random & $0.84 \pm 0.06$ & $0.82 \pm 0.05$ & $0.87 \pm 0.02$ \\
            Ours & $0.87 \pm 0.05$ & $0.90 \pm 0.03$ & $0.91 \pm 0.04$ \\
            Complete & $0.90 \pm 0.06$ & $0.91 \pm 0.03$ & $0.90 \pm 0.03$ \\ \hline
        \end{tabular}
    }
    \caption{Test classification accuracy on SpokenArabicDigits (mean ± std over 4 seeds, \%).}
    \label{tab:different-fakes-spoken}
\end{table}

Our acquirer consistently outperforms the random acquisition policy, and often even matches the performance of the complete acquirer. The acquisition patterns show that our acquire starts selecting the real features (notice the horizontal lines), although still occasionally sampling fake features. Additionally, we note that in some cases, the complete acquirer exhibits overfitting, while our acquirer avoids it (e.g., for noise fake features on m-FordA, shown in \cref{fig:training-curves}).

\subsection{Shifted fake features}

\begin{figure}[h]
    \centering
    \subfigure[Adding fake features]{
        \includegraphics[page=4, trim={0 2.1in 3.76in 0}, clip, width=0.24\linewidth]{images/figures.pdf}
        \label{fig:normal-fake}
    }\hfill
    \subfigure[Random acquirer]{
        \includegraphics[width=0.24\linewidth]{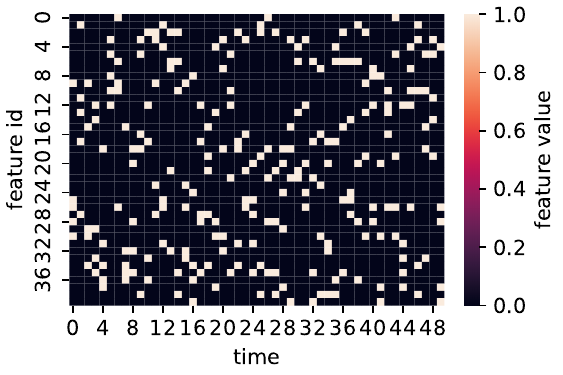}
    }\hfill
    \subfigure[Our acquirer]{
        \includegraphics[width=0.24\linewidth]{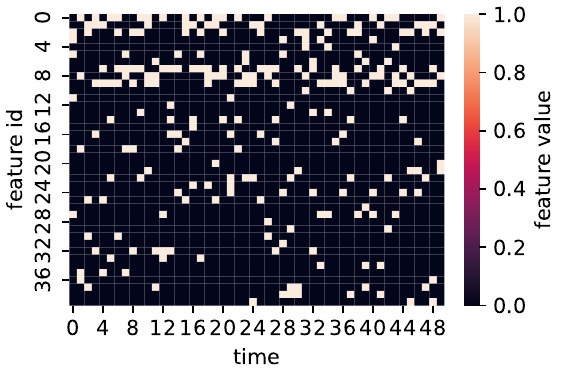}
    }
    \medskip\\
    \subfigure[Shifting fake features]{
        \includegraphics[page=5, trim={0 2.1in 3.76in 0}, clip, width=0.24\linewidth]{images/figures.pdf}
        \label{fig:shifted-fake}
    }\hfill
    \subfigure[Static acquirer]{
        \includegraphics[width=0.24\linewidth]{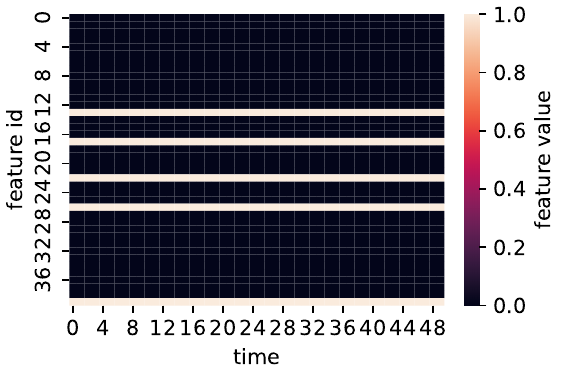}
    }\hfill
    \subfigure[Our acquirer]{
        \includegraphics[width=0.24\linewidth]{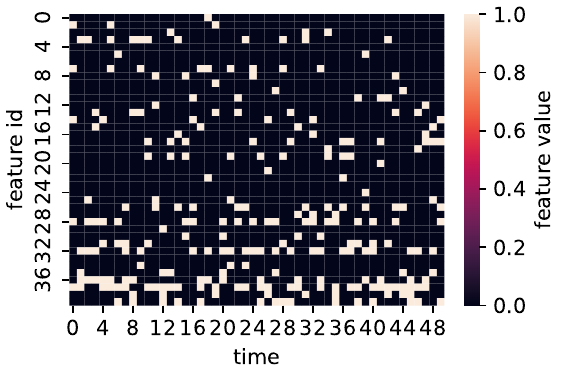}
    }
    \caption{Data modification sketch and acquisition patterns on m-FordA (by row).}
    \label{fig:fake-features}
    \label{fig:acq-patterns-forda}
\end{figure}

To test whether the learned acquisition is dynamic, we shift the real features so that the acquisition pattern would have to change over time (see \cref{fig:shifted-fake}). The dynamic policy should be able to learn that shift, while a static acquirer will only select the same set of features throughout the time series. We use a random forest (RF) as a static feature selection baseline, as it has been  used for feature importance analysis of ICU data \citep{hyland_early_2020}.

The results and the acquisition patterns are shown in \cref{tab:fake-features-shifted-forda,fig:acq-patterns-forda}. Our acquirer outperforms the random policy, but is outperformed by the static policy. The acquisition pattern shows that the model does not manage to capture the shift in fake features. 

We hypothesize that the underperformance of our acquirer is due to its simplistic architecture. The time step is passed to the model, but it does not receive the hidden state of the classifier. A more sophisticated architecture (e.g., an LSTM) that receives the classifier state as input will likely perform better.

\section{Conclusion}

Dynamic feature acquisition is a challenging problem that arises for temporal data across various applications: medicine, wearable sensors, active perception, etc. It has seen little attention, with previous work considering only reinforcement learning approaches.

In this work, we propose to dynamically select the most informative features using an approach similar to CMI maximization. We show that the acquirer trained using our approach learns to distinguish fake features from real ones for time series classification. Our model outperformed a random acquisition policy, but it did not surpass the static acquisition. This performance gap is likely due to the simplicity of the used architectures. 

We hope that this work will be continued, as a wide range of questions remain open. Future work may consider more advanced architectures, compare the performance of our training approach to reinforcement learning \citep{kossen_active_2023}, and loosen the assumptions we adopted: fixed time step budget, fully observed training data, and equal feature acquisition cost.

\section*{Acknowledgements}

FS thanks Shkurta Gashi and Manuel Burger for helpful discussions. Computational data analysis was performed at \href{https://sis.id.ethz.ch/services/sensitiveresearchdata/}{Leonhard Med} secure trusted research environment at ETH Zurich. FS was supported by grant \#902 of the Strategic Focus Area “Personalized Health and Related Technologies (PHRT)” of the ETH Domain (Swiss Federal Institutes of Technology). VF was supported by a Branco Weiss Fellowship.

\bibliography{feat-acq}

%%%%%%%%%%%%%%%%%%%%%%%%%%%%%%%%%%%%%%%%%%%%%%%%%%%%%%%%%%%%%%%%%%%%%%%%%%%%%%%
%%%%%%%%%%%%%%%%%%%%%%%%%%%%%%%%%%%%%%%%%%%%%%%%%%%%%%%%%%%%%%%%%%%%%%%%%%%%%%%
% APPENDIX
%%%%%%%%%%%%%%%%%%%%%%%%%%%%%%%%%%%%%%%%%%%%%%%%%%%%%%%%%%%%%%%%%%%%%%%%%%%%%%%
%%%%%%%%%%%%%%%%%%%%%%%%%%%%%%%%%%%%%%%%%%%%%%%%%%%%%%%%%%%%%%%%%%%%%%%%%%%%%%%

\newpage
\clearpage
\appendix

\counterwithin{figure}{section}
\counterwithin{equation}{section}
\counterwithin{table}{section}

\section{Related work}
\label{sec:related}

\begin{figure}[htb]
    \centering
    \subfigure[Active learning]{
        \includegraphics[page=6, trim={0 6.36cm 11.5cm 0},clip, width=0.32\linewidth]{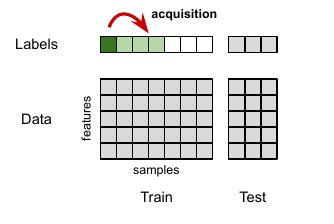}
    }%
    \hfill
    \subfigure[Active feature acquisition]{
        \includegraphics[page=7, trim={0 6.36cm 11.5cm 0},clip, width=0.32\linewidth]{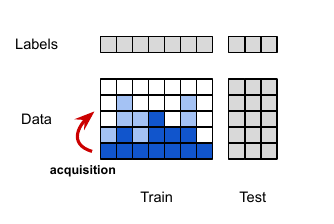}
    }%
    \hfill
    \subfigure[Dynamic feature acquisition]{
        \includegraphics[page=8, trim={0 6.36cm 11.5cm 0},clip, width=0.32\linewidth]{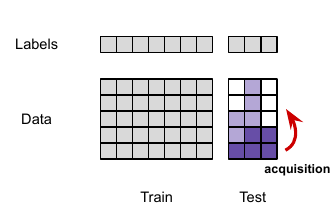}
    }
    \caption{Comparison of active learning, active and dynamic feature acquisition tasks on static data.}
    \label{fig:active-learning}
\end{figure}

To the best of our knowledge, the only prior work that has considered DFA on time series data is \citet{kossen_active_2023}. They use reinforcement learning and focus on multimodal data. Feature acquisition on static data has received wider attention. Both methods using mutual information \citep{ma_eddi_2019, lewis_accurate_2021, covert_learning_2023} and reinforcement learning \citep{yu_deep_2023} have been developed.

In \citet{ma_eddi_2019}, CMI is estimated by training a partial variational autoencoder (P-VAE). This allows the model to perform imputation from any subset of observed features and select the features associated with high-value information. In \citet{lewis_accurate_2021}, this approach has been developed further with the use of transformers for processing sets of observed features. The main challenge with using the P-VAE is its training. Training generative models can be challenging, especially for more complex data such as images \citep{covert_learning_2023}.

An alternative approach presented in \citet{covert_learning_2023} aims to approximate CMI instead of estimating it precisely. They propose using a Categorical distribution and greedily select the feature with the largest CMI at each step. Unlike \citet{ma_eddi_2019}, they use only simple (dense) architectures. However, their approach accepts set-based models as well.

For static ICU data, deep reinforcement learning has been used for DFA training \citep{yu_deep_2023}. The authors took into account that medical tests are usually done in panels (i.e., provide multiple features at the same time) and differ in cost. They also produce the accuracy-cost Pareto fronts, which help analyze the trade-off made when setting a specific acquisition budget.

DFA using CMI is closely related to active learning and active feature acquisition (see \cref{fig:active-learning}). Recent works show that Bayesian models can perform well in active learning \citep{sharma2023incorporating}. It has been shown that Bayesian acquisition functions such as Bayesian active learning by disagreement (BALD) are connected to CMI \citep{ma_eddi_2019}. Perhaps other Bayesian acquisition functions, such as expected predictive information gain \citep[EPIG;][]{smith2023prediction}, could be adapted for use in DFA.

For ICU time series data, feature importance has been studied using random forests \cite{hyland_early_2020}. In \citet{hyland_early_2020, yeche_temporal_2023} authors showed that deep learning architectures can achieve state-of-the-art performance in early event prediction. Tokenization of observed ICU features has been shown to improve the performance of such models \citep{kuznetsova_importance_2023}. Tokenization of observed features is a natural part of the set-based approaches \citep{ma_eddi_2019}, and could be applied in DFA.

\section{DFA}
\label{sec:assumptions}

\newcommand\mycommfont[1]{\footnotesize\ttfamily{#1}}
\SetCommentSty{mycommfont}
\setlength{\algomargin}{2em}

\begin{algorithm2e}
    \caption{Next-step DFA on regular time series}
    \label{alg:dfa}
    \DontPrintSemicolon
    \SetKwInOut{Input}{Input}
    \Input{time series $\pmb{x}$ of length $T$, time step budget function $b(\cdot,\cdot)$,\\acquirer with hidden state $h_t$, classifier with hidden state $H_t$}
    $t \gets 0$ \\
    $\pmb{m}_0 \gets$ acquirer.init() \tcp*{initial acquisition request}
    \While{$t < T$}{
        $\pmb{x}'_t \gets \pmb{m}_t \cdot \pmb{x}_t$ \tcp*{measure requested features}
        classifier.step$(\pmb{x}'_t, \pmb{m}_t, h_t, t)$ \\
        $\pmb{m}_{t+1} \gets$ acquirer.step$(\pmb{x}'_t, \pmb{m}_t, b(\pmb{x}_t,t), H_t, t$) \\
        $t \gets t + 1$ \\
    }
    $y_{pred} \gets$ classifier.predict() \tcp*{make the prediction}
    $C \gets \sum_{t=0}^{T} \pmb{m}_t$ \tcp*{calculate the cost}
\end{algorithm2e}

The ``next-step'' prediction assumption is satisfied when the time it takes to measure requested features is smaller than the time step duration. Both the ``next-step'' and regularity assumptions are plausible for ICU when a bigger resolution (e.g., one hour) is chosen \citep{hyland_early_2020}.

The assumptions about equal feature acquisition cost, fully observed data, and constant a prior set budget per time step do not hold for medical data. We leave generalization to future work.

\section{Datasets}
\label{sec:data}

\begin{table}[H]
    \centering
    \resizebox{\textwidth}{!}{%
        \begin{tabular}{lllllllll}
            \hline
            Dataset            & Task           & Classses & Domain             & Train size & Test size & Number of features & Length & Class balance \\ \hline
            FordA          & Classification & 2 & Sensor & 3601 & 1320 & 1      & 500 & Balanced \\
            m-FordA (m=10) & Classification & 2 & Sensor & 3601 & 1320 & 10 (m) & 50 & Balanced \\
            SpokenArabicDigits & Classification & 10       & Speach recognition & 6600       & 2200      & 13                 & 4-93   & Balanced      \\ \hline
        \end{tabular}%
    }
    \caption{Summary of the datasets.}
    \label{tab:data-stats}
\end{table}

\begin{figure}[htbp]
    \centering
    \subfigure[FordA]{
        \includegraphics[width=0.31\textwidth]{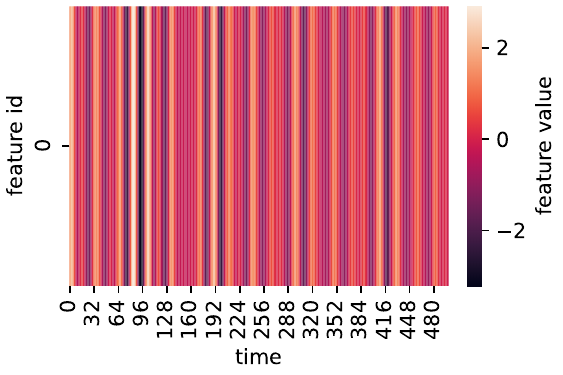}
    }
    \hfill
    \subfigure[m-FordA]{
        \includegraphics[width=0.31\textwidth]{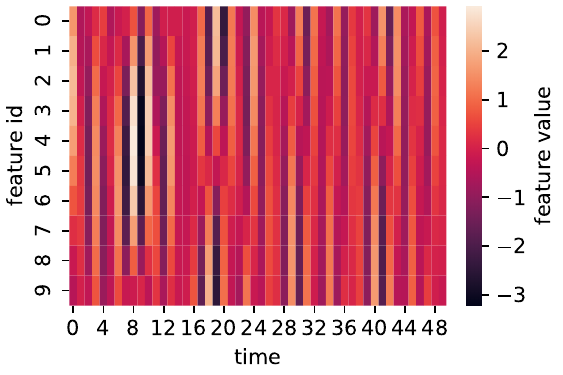}
    }
    \hfill
    \subfigure[SpokenArabicDigits]{
        \includegraphics[width=0.31\textwidth]{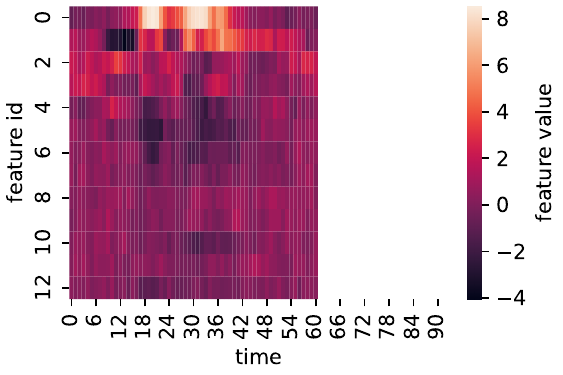}
    }
    \caption{Samples from the datasets.}
    \label{fig:data-samples}
\end{figure}

\newpage
The fake features are either zeros, sampled from Gaussian noise ($0$ mean and $0.5$ standard deviation), or sampled from a GP with an RBF kernel using \citet{scikit-learn} (amplitude coefficient $0.5$, length scale $1.5$, length scale bounds $[0.1,10]$). These parameters were selected so that the fake features are visually similar to real ones.

\begin{figure}[htbp]
    \centering
    \subfigure[Zeros]{
        \includegraphics[width=0.31\textwidth]{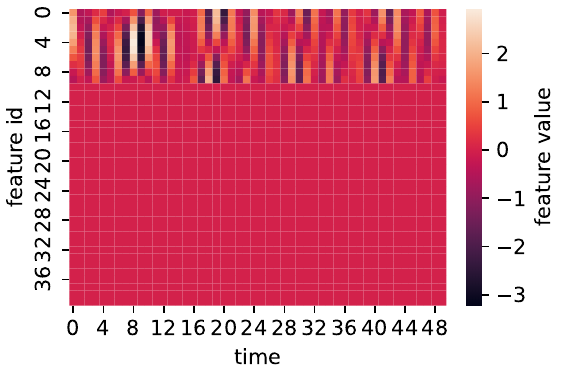}
    }
    \hfill
    \subfigure[Gaussian noise]{
        \includegraphics[width=0.31\textwidth]{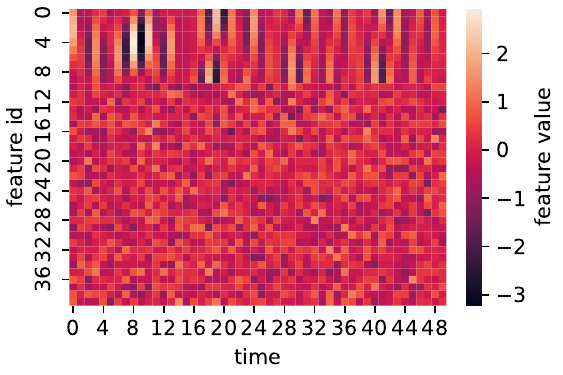}
    }
    \hfill
    \subfigure[Samples from a GP]{
        \includegraphics[width=0.31\textwidth]{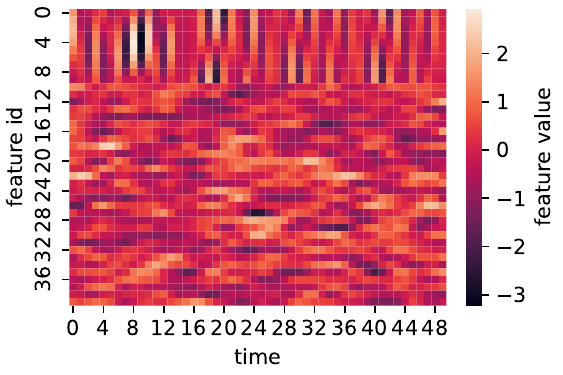}
    }
    \caption{A sample from the m-FordA dataset with $30$ fake features of different kinds.}
    \label{fig:data-ford-fakes}
\end{figure}

\begin{figure}[htbp]
    \centering
    \subfigure[Zeros]{
        \includegraphics[width=0.31\textwidth]{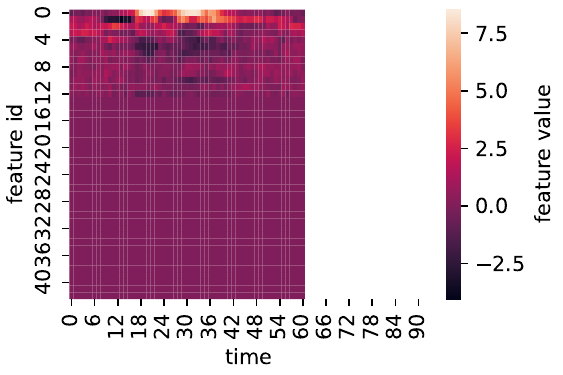}
    }
    \hfill
    \subfigure[Gaussian noise]{
        \includegraphics[width=0.31\textwidth]{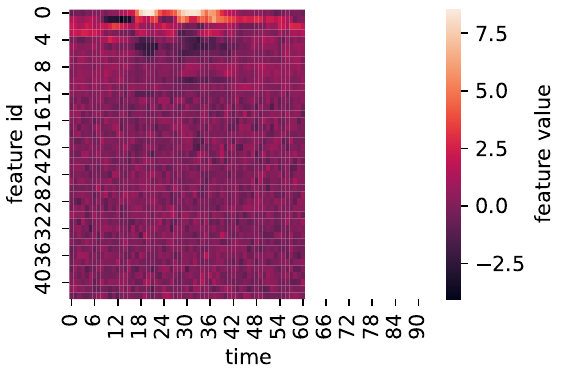}
    }
    \hfill
    \subfigure[Samples from a GP]{
        \includegraphics[width=0.31\textwidth]{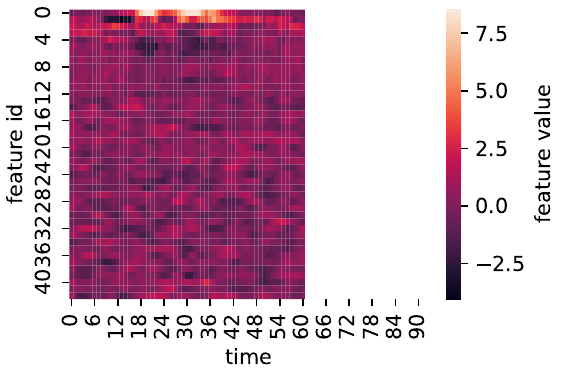}
    }
    \caption{A sample from the SpokenArabicDigits dataset with $30$ fake features of different kinds.}
    \label{fig:data-spoken-fakes}
\end{figure}

To create the data with shifted real features, they were swapped with fake features (by ids) every few time steps proportionally to their number. More specifically, if the number of the real features is $R$ and the number of fake features is $F$, the indices $i$ of real features will shift to $i + R$ every $\floor*{\frac{R}{R + F}} \cdot T$ time steps. For example, for m-FordA with $m=10$ with $20$ fake features, the real features will have indices $0$ to $10$ during the first third of the time steps, $10$ to $20$ during the second third, and $20$ to $30$ for the rest of the series (see \cref{fig:shifted-fake}). 

\begin{figure}[htbp]
    \centering
    \includegraphics[width=0.31\textwidth]{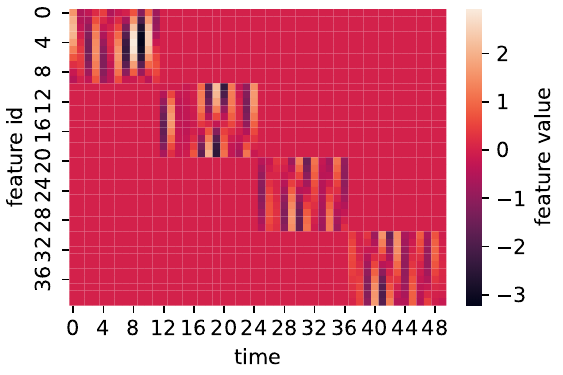}
    \caption{A sample from the m-FordA dataset with $30$ shifted fake features (zeros).}
    \label{fig:data-shifted-fakes}
\end{figure}

\section{Experiments}

\subsection{Setup details}
\label{sec:appendix}

The acquirers are implemented using fully connected neural networks with $1$ hidden layer ($4$ hidden units for m-FordA, and $8$ for SpokenArabicDigits) and ReLU activations. The input is formed by concatenating the previous observation, acquisition mask, and current time step. The internal classifier state is not passed to the acquirer. 

The classifiers are implemented using Long Short-Term Memory networks \citep[LSTMs;][]{lstm} with $16$ hidden units, $2$ layers for m-FordA and $3$ for SpokenArabicDigits (with ReLU activations), and a linear dimension of $8$, followed by one linear layer outputting class logits. 

We use a simpler training procedure compared to \citet{covert_learning_2023}: the temperature in the Gumbel distribution is fixed, and we do not pre-train the classifiers.  For logits vector $l$, the penalty function $R$ is 
$R(\pmb{l}) = 100 \cdot \pmb{m}_t \cdot |\pmb{l}|$, where the absolute value is taken elementwise.

We use a random forest (static feature selector) with $1000$ esimators, leaving the other parameters as defaults provided by scikit-learn \citep{scikit-learn}. 

We train using the Adam optimizer \cite{adam} with cross-entropy loss in PyTorch \cite{paszke2017automatic}. The batch size is $1000$, and the learning rate is $0.001$ in all experiments.

\subsection{Additional results}

\begin{table}[htbp]
    \centering
    \resizebox{0.475\textwidth}{!}{%
        \begin{tabular}{llccc}
            \hline
            \multirow{2}{*}{Acquirer} & \multicolumn{3}{c}{Type of fake features} \\ \cline{2-4}
            & \multicolumn{1}{c}{Zeros} & \multicolumn{1}{c}{Noise} & GP \\ \hline
            Random & $0.76 \pm 0.04$ & $0.74 \pm 0.04$ & $0.73 \pm 0.02$ \\
            Ours & $0.86 \pm 0.03$ & $0.86 \pm 0.03$ & $0.84 \pm 0.02$ \\
            Complete & $0.93 \pm 0.00$ & $0.84 \pm 0.02$ & $0.80 \pm 0.03$ \\ \hline
        \end{tabular}
    }
    \caption{Test classification accuracy on m-FordA (mean ± standard deviation over 5 seeds, \%).}
    \label{tab:different-fakes-forda}
\end{table}

\begin{table}[htb]
    \centering
    \resizebox{0.25\textwidth}{!}{%
        \begin{tabular}{lc}
            \hline
            Acquirer     & Accuracy, \%\\ \hline
            Random       & $0.708$ \\ 
            Static (RF)  & $0.842$ \\ 
            Ours         & $0.740$ \\ 
            Complete     & $0.897$ \\ \hline
        \end{tabular}
    }
    \caption{Test classification accuracy on m-FordA with fake features (zeros).}
    \label{tab:fake-features-shifted-forda}
\end{table}

\begin{figure}[htbp]
    \centering
    \subfigure[Complete acquirer]{
        \includegraphics[page=9, trim={0 5.08cm 8.57cm 0},clip, width=0.375\linewidth]{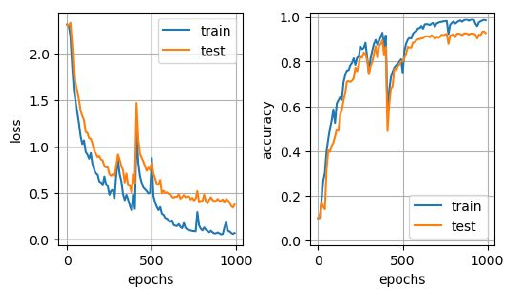}
    }
    \qquad
    \subfigure[Our acquirer]{
        \includegraphics[page=10, trim={0 5.08cm 8.57cm 0},clip, width=0.375\linewidth]{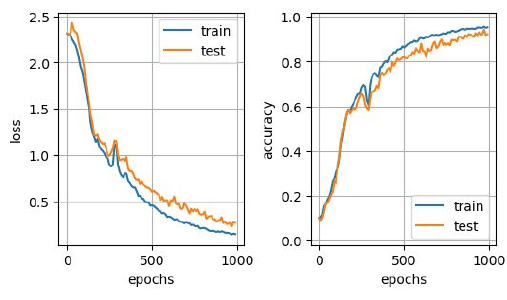}
    }
    \caption{Training curves on m-FordA with fake features (zeros).}
    \label{fig:training-curves}
\end{figure} 
\vspace{-3cm} % remove extra page

\end{document}